\documentclass[journal]{IEEEtran}

\usepackage{setspace}

\usepackage{graphicx}
\usepackage{textcomp}
\usepackage{float}
\usepackage[table,xcdraw]{xcolor}
\usepackage{CJKutf8}
\usepackage{setspace}
\usepackage{booktabs}
\usepackage{amsfonts}
\usepackage{amsthm,amsmath}
\usepackage{mathrsfs}
\usepackage{multirow}
\usepackage[noend]{algpseudocode}
\usepackage{algpseudocode}
\usepackage{diagbox}
\usepackage{color}
\usepackage{colortbl} 
\usepackage{xcolor}
\usepackage{array} 
\usepackage{bbding} 

\begin{document}
	
	\title{ECEA: Extensible Co-Existing Attention for Few-Shot Object Detection}

	\author{Zhimeng Xin,~\IEEEmembership{}
		Tianxu Wu,~\IEEEmembership{}
		Shiming Chen,~\IEEEmembership{}
		Yixiong Zou,~\IEEEmembership{}	
		Ling Shao,~\IEEEmembership{Fellow IEEE}
		and Xinge You,~\IEEEmembership{Senior Member,~IEEE }
		\thanks{		
			Z. Xin is with the School of Cyber Science and Engineering, Huazhong University of Science and Technology, Wuhan 430074, China (e-mail: zhimengxin@hust.edu.cn). 
			
			T. Wu, S. Chen and X. You are with the School of Electronic Information and Communications, Huazhong University of Science and Technology, Wuhan 430074, China (e-mail: wutianxu@hust.edu.cn; gchenshiming@gmail.com; and youxg@hust.edu.cn). 
			
			Y. Zou is with the School of Computer Science \& Technology, Huazhong University of Science and Technology, Wuhan 430074, China (e-mail:yixiongz@hust.edu.cn).
			
			L. Shao is with UCAS-Terminus AI Lab, UCAS. (e-mail: ling.shao@ieee.org)
			
			(\textit{Corresponding author: Shiming Chen; Co-First Author: Zhimeng Xin and Tianxu Wu})
		}
	}

	\maketitle

	\begin{abstract}
		Few-shot object detection (FSOD) identifies objects from extremely few annotated samples. Most existing FSOD methods, recently, apply the two-stage learning paradigm, which transfers the knowledge learned from abundant base classes to assist the few-shot detectors by learning the global features. However, such existing FSOD approaches seldom consider the localization of objects from local to global. Limited by the scarce training data in FSOD, the training samples of novel classes typically capture part of objects, resulting in such FSOD methods cannot detect the completely unseen object during testing. To tackle this problem, we propose an Extensible Co-Existing Attention (ECEA) module to enable the model to infer the global object according to the local parts. Essentially, the proposed module continuously learns the extensible ability on the base stage with abundant samples and transfers it to the novel stage, which can assist the few-shot model to quickly adapt in extending local regions to co-existing regions. Specifically, we first devise an extensible attention mechanism that starts with a local region and extends attention to co-existing regions that are similar and adjacent to the given local region. We then implement the extensible attention mechanism in different feature scales to progressively discover the full object in various receptive fields. Extensive experiments on the PASCAL VOC and COCO datasets show that our ECEA module can assist the few-shot detector to completely predict the object despite some regions failing to appear in the training samples and achieve the new state of the art compared with existing FSOD methods.
	\end{abstract}
	

	\IEEEpeerreviewmaketitle

	\section{Introduction}
	
	\begin{figure}[t]
		\vspace{0.1cm}
		\centering
		\includegraphics[width=9cm,height=10.71cm]{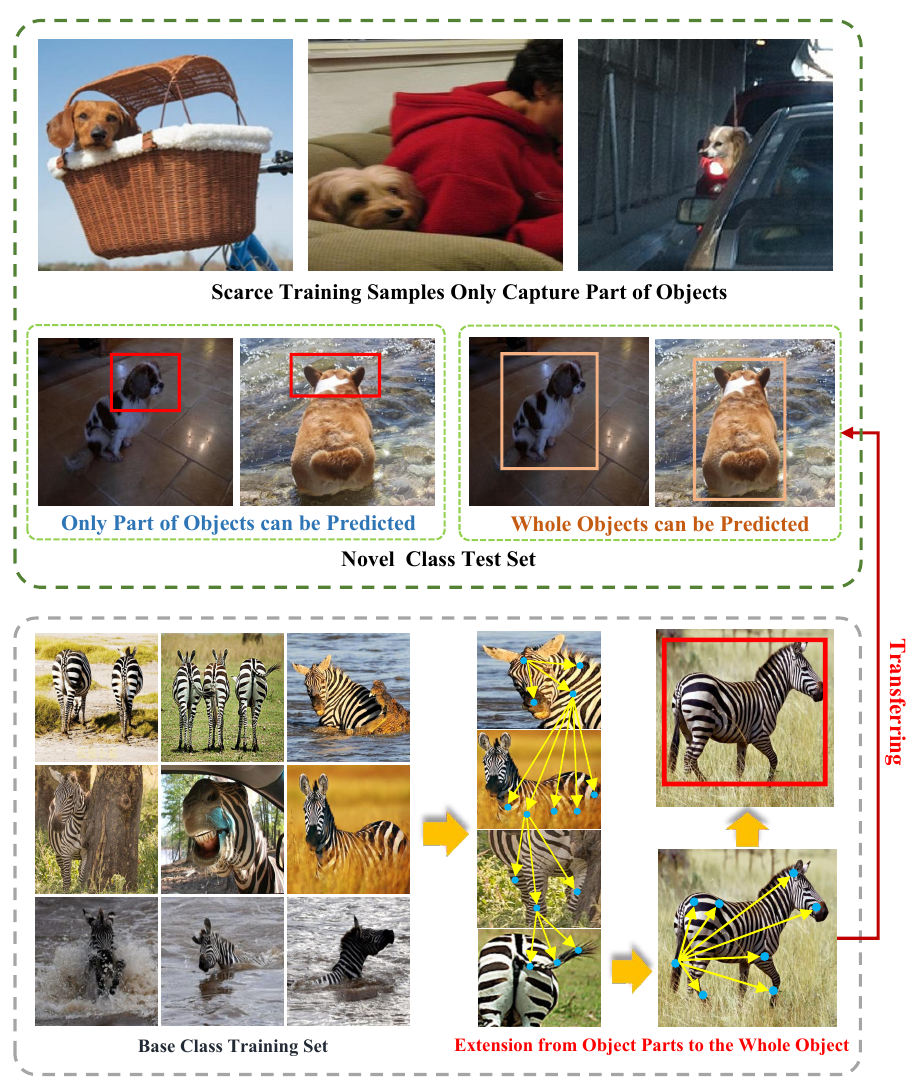} 
		\vspace{-0.3cm}
		
		\caption{Motivation illustration. Limited by scarce training data, novel classes are highly possible to provide training samples that only capture part of objects, making it hard for the existing FSOD models to detect the complete objects during testing. We consider that base classes contain adequate training samples to capture the complete semantic object, even though each training sample may contain only part of the objects. Therefore, we propose an extension mechanism to gradually associate object parts to infer the complete object and transfer the extensible ability from base classes to novel classes to infer the co-existing features that are unseen object parts in the training stage.
		}

		\label{f1}
		\vspace{-0.1cm}
	\end{figure}

	\IEEEPARstart{F}{ew-shot} object detection (FSOD) pertinent to object detection has recently emerged as an area of particular interest \cite{fslsurvey}. The motivation of such a line of research is that FSOD improves the performance of object detection in few-shot annotated samples since it combines object detection \cite{yolov2, fasterrcnn,tipob1,transob2} and few-shot learning \cite{fsl1,fsl2,fsl3,fsl4,tip3,tip4} to rapidly adapt novel concepts. These studies may complement previous studies and reduce the limitation of the over-reliance on large-scale annotated samples \cite{sfsod2,sfsod3,sfsod}. 
	
	Most existing FSOD methods, recently, apply the two-stage learning paradigm, which transfers the knowledge learned from abundant base classes to assist the few-shot detectors by learning the global features.  Such methods could be divided into meta-learning-based approaches \cite{fsrw, fct, vfa} and transfer-learning-based methods \cite{TFA, fsce, defrcn}. Meta-learners learn a set of initialization parameters to improve performance by serving each task as a unit in iterative training. The transfer-learning-based methods freeze the pre-trained parameters to only fine-tune the last layers of the detector in the novel stage to reach or outperform the existing meta-based strategies.

	However, such FSOD approaches \cite{defrcn, vfa} seldom consider the localization of objects from local to global. Limited by the scarce training data in FSOD, novel classes are highly possible to provide training samples that only capture part of objects, resulting in the existing FSOD methods \cite{TFA, fct,defrcn, vfa} cannot detect the complete object during testing. As shown in Fig. \ref{f1} (top), only the head of the dog can be seen in three novel class training samples, but the model is required to predict the dog's body and tail which are unseen in the scarce training samples. This inevitably causes the model to excessively focus on the head of the dog and ignore the unseen object parts, resulting in only the dog's head can be detected for test samples. As such, properly designing effective methods to tackle such problems is very necessary for advancing FSOD.

	In this paper, we consider transferring information from base classes to novel classes. Unlike novel classes, base classes contain adequate training samples to capture the complete semantic object, even though each training sample may contain only part of the objects. As shown in Fig. \ref{f1}, base classes contain a large number of training samples for zebra. Although each sample may only capture parts of the zebra, all training samples together represent the whole zebra, if the model with extensible ability can effectively detect the complete zebra during testing. Extensible models trained with sufficient data have such ability because the model learns to associate object parts that co-exist within the same semantic class. Therefore, even though some parts of the object are hard to be detected during testing, the Extensible model can extend the easily detected region to co-existing regions to detect the complete object, as shown in Fig. \ref{f1} (bottom).

\begin{figure*}[t]
		\vspace{-0.1cm}
		\centering
		\includegraphics[width=17cm,height=11.8cm]{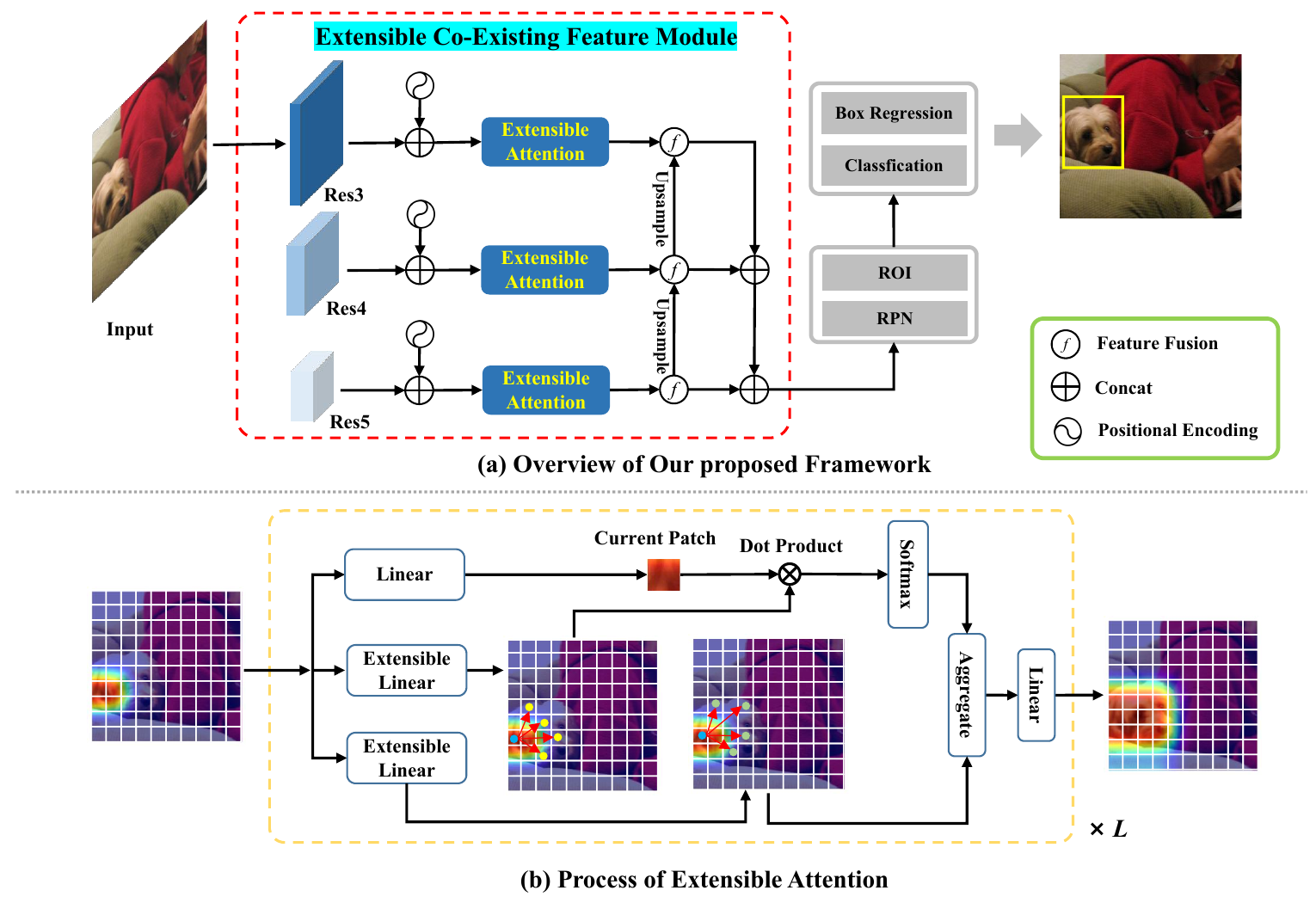} 
		\vspace{-0.1cm}
		\caption{The framework of our proposed ECEA. (a) The output of the last three stages (Res3, Res4, and Res5) from ResNet101 \cite{res101}, combined with positional encoding are separately input to the extensible attention module. Then the output of extensible attention fuses with the adjacent previous extensible attention stage by upsample. We finally concatenate the output of three stages input into the anchor generation layers. (b) Extensible attention splits the feature map into a series of patches and extends from a certain patch to adjacent relevant patches by extensible linear layers. In the two feature maps following the extensible Linear layer, the blue circle represents a certain point, while the yellow and green circles represent the extensible points. \textit{L} is the number of extensible attention layer.}
		\label{f2}
	\end{figure*}

	In light of these observations, we propose an \textbf{E}xtensible \textbf{C}o-\textbf{E}xisting \textbf{A}ttention (\textbf{ECEA}) module to enable the model to infer the global object according to the local parts. Essentially, the model continuously learns the extensible ability on the base stage with abundant samples and transfers it to the novel stage, which can assist the few-shot model to quickly adapt in extending local regions to co-existing regions. Specifically, we first design the extensible attention mechanism that starts with a local region and extends attention to co-existing regions that are similar and adjacent to the given local region. The extension operation is repeated on the extended regions until all co-existing regions are covered. We then implement the extensible attention mechanism in different feature scales to progressively discover the full object in various receptive fields, as shown in Fig. \ref{f2}(a). Extensive experiments on the PASCAL VOC and COCO datasets show that the ECEA module can assist the few-shot detector to completely predict the object despite some regions failing to appear in the training samples and greatly improve the performance of few-shot detectors.

	Our main contributions can be summarized as follows: 
	
	\begin{itemize} 
		
		\item We propose an Extensible Co-Existing Attention (ECEA) module, which assists the model to infer co-existing unseen features according to the provided local part.

		\item We design an extensible attention mechanism to extend co-existing regions and implement it in different feature scales to progressively discover the full object in various receptive fields.
		
		\item Extensive experiments on the PASCAL VOC and COCO datasets demonstrate that our method achieves the new state of the art (SOAT). We further qualitatively show that our ECEA module can assist the few-shot detector to completely predict the object despite some regions failing to appear in the training samples.

	\end{itemize}

	The remainder of this article is organized as follows. Related works are discussed in Section II. The proposed ECEA module is illustrated in Section III. Experimental results and analysis are provided in Section IV. Finally, we summarize the discussions and conclusions in Section V and VI, respectively.

	\section{Related Work}

	\subsection{General Object Detection}
	
	Extensive works have investigated mainstream object detection methods \cite{flob,acmob,tipob1,tipob2} from different perspectives that could be grouped into two categories of one-stage detectors and two-stage. In general, one-stage-based object detection methods, e.g., YOLO ~\cite{yolov1,yolov2} and SSD~\cite{ssd}, only need to predict the anchors in a batch of images once before the regression and classification layer to achieve ideal performance. In contrast, two-stage object detectors, e.g., Faster R-CNN~\cite{fasterrcnn} and Fast R-CNN~\cite{Fastrcnn}, first generate a series of anchors by using region proposal network (RPN) from the feature images and then perform region proposals classification and regression. Such two-stage detectors supplement RPN modules which negatively impact detection speed but tend to have higher detection performance compared with the former methods~\cite{yolov2,ssd}. However, these mainstream researches rely on abundant annotated data, which may not bring a positive influence in data-scarce scenarios.

	\subsection{Few-Shot Object Detection}
	
	To solve the negative detection effects of scarce annotated samples, FSOD methods have increasingly received attention. They are roughly categorized into two types, namely, \cite{fsrw,metadet,qa,metafrcn,metarcnn,fct,vfa,tip2} being the meta-learning-based methods and \cite{TFA,fsce,defrcn,cfa,sfsod3} being the transfer-learning-based methods.

	As for meta-learning-based methods, FSRW~\cite{fsrw} introduces a meta-feature learner and a lightweight feature reweighting module to reorganize the detector which enables the model to quickly adapt to novel categories. Different from FSRW, Meta R-CNN~\cite{metarcnn} only uses the meta-learning paradigm in the region of interest (ROI) features which are fused with the support branch to obtain the category attention vectors. Xiao\textit{ et al}.~\cite{xiao} further improved the fusion network based on Meta R-CNN to obtain better detection performance. Furthermore, compared with the above CNN-based methods, some transformer-based meta-learning FSOD methods achieve better performance. For instance, Zhang \textit{et al}.~\cite{metadetr} proposed Meta-DETR based on Deformable DETR~\cite{detr}, which combines with transformer and meta-learning to ease the dependence on RoI. Han \textit{et al}.~\cite{fct} proposed a Fully Cross-Transformer based model to optimize the few-shot similarity learning between the two branches.

	Concerning transfer-learning-based FSOD, TFA uses cosine similarity to measure the relevance between the candidate anchors and the ground truth bounding boxes in the fine-tuning stage. There have many followed-up kinds of research based on this simple and effective method. Sun \textit{et al}.~\cite{fsce} considered that the candidate boxes with different scores of Intersection over Union (IoU)~\cite{IOU} are similar to intra-class data augmentation. Based on this fact, they proposed FSCE which doubles the maximum number of candidate boxes processed by NMS~\cite{NMS} and halves the number of boxes used for loss calculation in ROI. According to the existing semantic relations between different categories, Zhu \textit{et al.}~\cite{ssr} proposed SRR-FSD based on TFA~\cite{TFA}, which integrates semantic relations augmented by relational reasoning into the fine-tuning stage. Qiao \textit{et al.}~\cite{defrcn} found that the entire FSOD model may suffer from being dominated by one of three components, including backbone, RPN, and RCNN, in the transfer learning-based detector, i.e., Faster R-CNN. They thus developed DeFRCN to alleviate the problem. Yang \textit{et al.} \cite{tip1} proposed an efficient pre-train-transfer framework to enhance the adaptation speed which determines the efficiency of the few-shot transfer process. 
	On the other hand, the directly fine-tuning approaches may lead to catastrophic forgetting of the base class in the training novel stage. To avoid forgetting, \cite{retina} and \cite{cfa} proposed Retentive R-CNN by utilizing consistency loss and a constraint-based method by adapting continual learning, respectively.
	Yet, these approaches seldom consider the localization of objects from local to global, i.e., neglect extended co-existing regions. Coupled with the limitation of the data-scarce scenario, some complex scenarios cannot be provided in the training stage, resulting in existing few-shot detectors incompletely predicting the box or even missing it in the test stage.

	\subsection{Extensible Learning}
	To capture the global features, some studies \cite{selfattention, scable} straightforwardly or indirectly put forward the concept of extensible learning. For example, Aditya \textit{et al.} \cite{scable} directly proposed scalable feature learning for nodes in graph networks, but it is difficult to extract features from the graph structure in the supervised tasks. On the contrary, the self-attention mechanism can capture all dependencies between image patches in a feature map to appropriately aggregate the input signal from the local to global in the supervised tasks. Inspired by this, some self-attention-based methods \cite{detr,ddetr,transob1,transob2} expand the receptive field to directly capture the complete regions of the object. Yet, these approaches rely on abundant annotated samples. To solve the problem, the self-attention mechanism has also made certain progress in FSOD tasks \cite{metadetr,fct}. For example, based on meta-learning, Han \textit{et al}. proposed FCT that aligns the extracted features in all network layers between the support set and query set to improve their similarity. Although FCT allows the model to acquire many features, it still cannot infer the unseen regions from the existing feature. In view of this, we propose an ECEA module that not only allows the model to learn the relationships between co-existing regions but infer the global object from part regions of the object in FSOD.
	
	\section{Method}
	
	In this section, we first introduce the formulation setting of FSOD in Section \ref{3.1}. We then propose an ECEA module to alleviate the data scarcity problem for FSOD by detecting the unseen object parts that are not provided in the few-shot training data in Section \ref{3.2}. We finally analyze the proposed model optimization in Section \ref{3.3}.

	\subsection{Formulation Setting}
	\label{3.1}
	
	\subsubsection{Problem Definition}
	
	We follow the present FSOD settings~\cite{TFA,fsce} in this paper. Denote  $D=\left\{ \left( x,y \right) ,x\in X,y\in Y \right\}$ as a dateset with a series of categories $C$, where $x$ represents the input image and $y=\left\{ c_i, b_i\right\}_{i=1}^{K}$ is the corresponding manual information with class $c$ and bounding box $b$. We then split $D$ into the sufficient annotated base dataset $D_b$ with its categories $C_b$ and only a few amounts of labeled novel dataset $D_n$ with its classes $C_n$ (usually less than 10), where $C_b \cup C_n=C$ and $C_b \cap C_n=\emptyset$.  
	
	As for the training, we adopt the two-stage fine-tuning paradigm. In the first stage, $D_b$ is used to train the initializing model $\mathcal{M}_{\text {init}}$ to obtain the base model $\mathcal{M}_{\text {base}}$. In the novel stage, only $D_n$ is utilized for training the FSOD model $\mathcal{M}_{fsod}$. Besides that, if the dataset is $C_b \cup C_n$ to compose a balanced dataset $D_f$ that only has a few annotated categories $C$, which is called generalized few-shot object detection (G-FSOD). Therefore, in this paper, the training process of FSOD or G-FSOD can be given by
	
	\begin{equation}				
	\mathcal{M}_{\text {init }} \stackrel{D_b}{\Longrightarrow} \mathcal{M}_{\text {base }} \stackrel{D_n / D_f}{\Longrightarrow} \mathcal{M}_{fsod } / \mathcal{M}_{g-fsod },
	\end{equation}				
	where $ \Longrightarrow $ indicates  the training process and $\mathcal{M}_{g-fsod }$ represents the G-FSOD model. With regard to evaluation, there is to calculate the results of each class, i.e., $C$, in G-FSOD~\cite{defrcn} and only to assess the results of $C_n$ in FSOD.

	\subsubsection{Transfer-Learning Based Framework}
	Faster R-CNN \cite{Fastrcnn} is a commonly used architecture with excellent performance in most transfer-learning-based FSOD methods~\cite{TFA, fsce, defrcn, cfa}. We also take Faster R-CNN as the basic architecture to demonstrate the performance of our proposed ECEA module, which assists the detector to extend the detection capability of features from local to global. As for a two-stage stacking framework, Faster R-CNN \cite{Fastrcnn} is composed of three independent functional modules which include backbone, RPN, and ROI. To be specific, the generalized features extracted by the backbone are respectively input into RPN used to generate class-independent proposals and ROI used to perform specific tasks related to class-related classification and localization.

	\subsection{Extensible Co-Existing Attention}
	\label{3.2}
	
	To enable the model to infer the global object according to the local parts in data-scarce scenarios, we propose a transfer-learning-based method, i.e., the ECEA module. Specifically, we develop extensible attention which lets the current object region from the input feature map extend to several co-existing regions and the obtained regions continue to extend outward. Then, we combine the feature extractor and the proposed extensible attention in the ECEA module as shown in Fig. \ref{f2}(a). Following previous works \cite{TFA,fsce}, we use ResNet101 \cite{res101} as the backbone (including five stages) for feature extraction. Experiments show that the combination of the last three stages with extensible attention can obtain the best performance in FSOD. Therefore, the output of the last three stages with the positional encoding is input into the proposed extensible attention that can learn the feature relationships in different feature scales to progressively discover the full object in various receptive fields. Furthermore, the outputs of attention layers in different scales are fused with each other through upsampling, which can help the high-level stage to correct the extensible unseen regions of former stages and the output of ECEA module to recover the high spatial resolution. Finally, we aggregate the fused stages to enrich the extensible feature information and promote the performance of the model in FSOD. The process of feature fusion can be given by
	
	\begin{equation}
	\begin{aligned}
	f_{Block4}&=\Psi_{up}\left( \varPhi_{Block5} \right) +\varPhi_{Block4}\\
	f_{Block4}&=\Psi_{up}\left( f_{Block4} \right) +\varPhi_{Block3}\\
	f_{output}&=\varPhi_{Block5}\oplus f_{Block4}\oplus f_{Block3} ,
	\end{aligned}
	\end{equation}
	where, $\varPhi_{Block}$ represents the output of Res combined with extensible attention, $\Psi_{up}(\cdot)$ represents upsampling, and $\oplus$ is matrix concat.

	\begin{figure}[t]
		\vspace{-0.1cm}
		\centering
		\includegraphics[width=8cm,height=8.96cm]{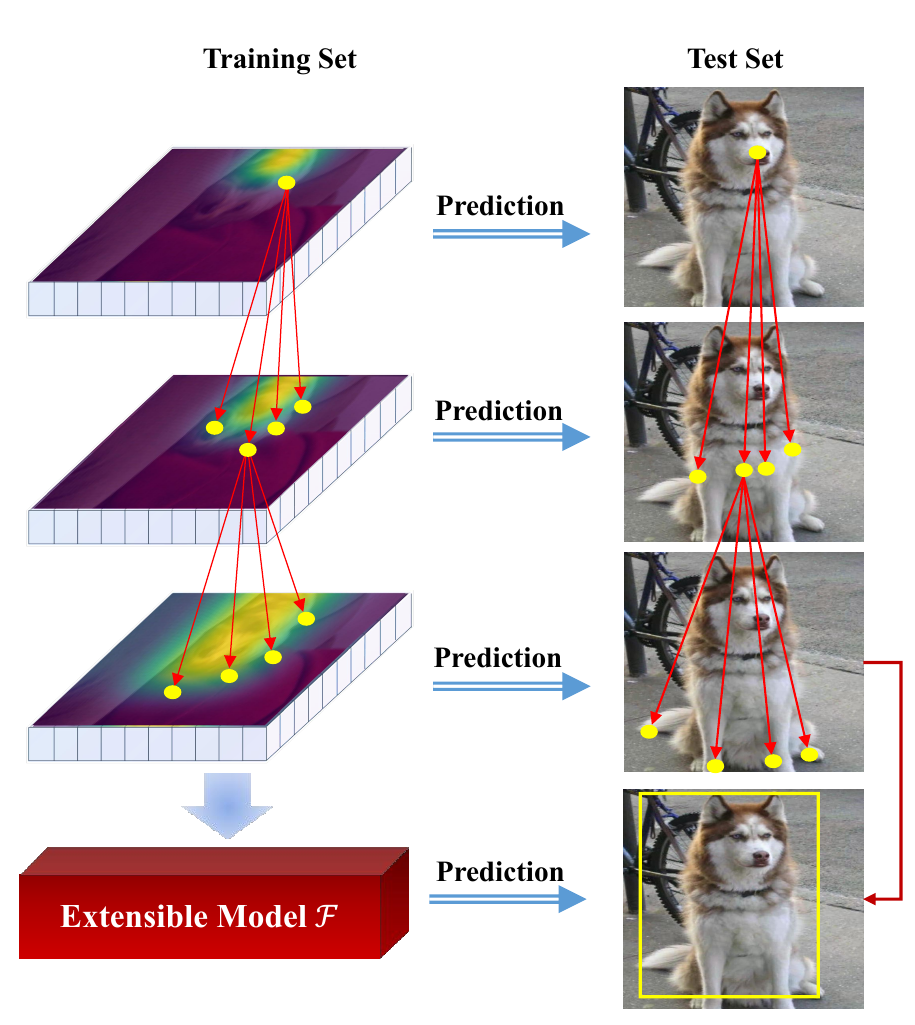} 
		\vspace{-0.1cm}
		\caption{Illustration of the extensible attention from support set to query. Each line represents once extensible attention layer.}
		
		\label{f3}
		\vspace{-0.1cm}
	\end{figure}
	
	\subsubsection{Extensible Attention}
	
	Fig. \ref{f2}(b) illustrates the processing of extensible learning. Give the input feature map $x \in R^{C \times H \times W}$ from the feature extractor, where $H$ and $W$ are the height and width of the map. In extensible attention, $x$ is divided into a series of region patches. A current patch from $x$ extends surrounding to learn the co-existing region patches. The obtained patches continue to learn the extensible region patches until the complete feature image is traversed. Furthermore, each region patch in the feature map is equivalent to the corresponding pixel point in $x$, one of which denotes $x_q$. And meanwhile, $N$ extensible region points of $x_q$ would be obtained by using the offset weight calculation \cite{ddetr,deforablercnn} in an extensible linear layer. 
	Denote $W^q$ as the linear layer to calculate the weight of $x_q$ corresponding to the current patch in Fig. \ref{f2}(b). Let $W_n^k$ and $W_n^v$ respectively represent the extensible linear layers to obtain key and value points of $x_q$. 
	the relevance of $x_q$ and key points is determined by the dot product layer. We use softmax to normalize the similarity score to aggregate the value points. 
	Therefore, once extensible attention calculation for $x_q$ can be given by

	\begin{equation}
	\operatorname{EA}\left(x_q\right)=\sum_{n=1}^N \frac{e^{x_q W^q \cdot\left(x_q W_n^k\right)^T}}{\sum_{n=1}^N e^{x_q W^q \cdot\left(x_q W_n^k\right)^T}} \cdot x_q W_n^v ,
	\end{equation}
	where $N$ is usually less than 10 in extensible attention. Thus, each $q$ only needs to pay attention to $N$ points.
	
	After $x_q$ extending, extensible attention does not directly calculate the attention of the obtained points by this location. It computes the attention of the point adjacent to $x_q$ until the entire feature map is traversed. The whole process belongs to parallel operation. This not only allows the model to extend the regions of a certain $x_q$, but also relates the whole feature map to $x_q$. The purpose is to make each feature point not only determine which feature points co-exist in the same object, but also which feature points are the background or other objects. By this, the computational complexity of $\operatorname{EA}\left(x\right)$ could be given by $\mathrm{O}\left(N_q C^2+N C^2+ N N_q \mathrm{C}\right)$,  where $N_q$ represents the number of the pixel points of $x$ and  $N_q >> N >> C$. Thus, the complexity can be simplified as $\mathrm{O}\left( N N_q \mathrm{C}\right)$ which is equivalent to $\mathrm{O}\left( N H W \mathrm{C}\right)$ in a feature map at the pixel level.

	\subsubsection{ Multi-Head and Multi-Layer Extensible Attention}
	To ensure the generalization ability of the model, we adopt the multi-head and multi-layer trick from transformer \cite{selfattention}. The multi-head extensible attention of an input image $x$ can be given by 
	
	\begin{equation}
	\operatorname{MHEA}(x)=\sum_{m=1}^M W_m \operatorname{E A}(x) ,
	\label{E2}
	\end{equation}
	where $W_m$ represents weight vectors aggregation and $M$ is the number of multi-head. 
	
	Increasing the extensible attention layer, the discriminant analysis of object features would be strengthened, which improve the robustness of object regression in the data-scarce scenario. 
	For example, Fig. \ref{f3} shows the model learning performance on the support and query samples in different extensible attention layers and respectively lists the learning performance of the model at each layer, where the output of the former layer is the input of the latter layer. Although the input image only contains part of the object information, with the deepening of the layers, the extensible ability of the model is gradually stimulated to completely cover the localization of the query object.
	Denoting $L$ as the number of multi-layer, thus multi-head and multi-layer extensible attention $\Phi(\cdot)$ can be expressed as:

	\begin{equation}
	\begin{gathered}
	\Phi(1)=\operatorname{MHEA}(x) \\
	\Phi(2)=\operatorname{MHEA}(\Phi(1)) \\
	\ldots \\
	\Phi(L)=\operatorname{MHEA}(\Phi(L-1)).
	\end{gathered}
	\end{equation}

	Furthermore, in a batch feature map, multi-layer ECEA enables the sample location of a certain map to indirectly learn several extensible co-existing regions from other feature maps and the obtained regions could correlate with co-existing regions from different images.

	\begin{table*}[t]
		\centering	
		\tabcolsep=0.15cm
		\caption{Performance comparison among ECEA and mainstream  methods based on PASCAL VOC with three random novel splits. The results are reported by averaging over multiple runs. Bold font indicates the SOTA result in the group.
		}
		\begin{tabular}{@{}ll|ccccc|ccccc|ccccc@{}}
			\toprule
			& \multirow{2}{*}{Methods / shots} &   \multicolumn{5}{c|}{Novel Split1}                                                                                    & \multicolumn{5}{c|}{Novel Split2}                                                                                    & \multicolumn{5}{c}{Novel Split3}                                                                                                         \\ 
			&	&                   \multicolumn{1}{c}{1}                         & \multicolumn{1}{c}{2}                         & \multicolumn{1}{c}{3}                         & \multicolumn{1}{c}{5}    & \multicolumn{1}{c|}{10}    & \multicolumn{1}{c}{1}                         & \multicolumn{1}{c}{2}                         & \multicolumn{1}{c}{3}                         & \multicolumn{1}{c}{5}    & \multicolumn{1}{c|}{10}     & \multicolumn{1}{c}{1}                         & \multicolumn{1}{c}{2}                         & \multicolumn{1}{c}{3}                         & \multicolumn{1}{c}{5}    & \multicolumn{1}{c}{10}                         \\ \midrule

			& \textit{FSOD} \\
			&	\multicolumn{1}{l|}{LSTD~\cite{lstd}}                  &  8.2                       & 1.0                       & 12.4                      & 29.1 & 38.5 & 11.4                      & 3.8                       & 5.0                       & 15.7 & 31.0 & 12.6                      & 8.5                       & 15.0                      & 27.3                      & 36.3                      \\ 

			&	\multicolumn{1}{l|}{TFA w/ cos~\cite{TFA}}                            &  39.8                      & 36.1                      & 44.7                      & 55.7 & 56.0 & 23.5                      & 26.9                      & 34.1                      & 35.1 & 39.1 & 30.8                      & 34.8                      & 42.8                      & 49.5                      & 49.8                      \\ 
			&	\multicolumn{1}{l|}{MetaDet~\cite{metadet}}                        &  18.9                      & 20.6                      & 30.2                      & 36.8 & 49.6 & 21.8                      & 23.1                      & 27.8                      & 31.7  & 43.0 & 20.6                      & 23.9                      & 29.4                      & 43.9                      & 44.1                      \\ 
			
			&	\multicolumn{1}{l|}{Meta R-CNN~\cite{metarcnn}}                     &  19.9                      & 25.5                      & 35.0                      & 45.7 & 51.5 & 10.4                      & 19.4                      & 29.6                      & 34.8 & 45.4 & 14.3                      & 18.2                      & 27.5                      & 41.2                      & 48.1                      \\

			&
			
			\multicolumn{1}{l|}{FSCE~\cite{fsce}}                           &  44.2                      & 43.8                      & 51.4                     & 61.9 & 63.4 & 27.3                      & 29.5                      & 43.5                      & 44.2 & 50.2 & 37.2                      & 41.9                      & 47.5                      & 54.6                      & 58.5                      \\ 
			
			&	QA-FewDet~\cite{qa} & 42.4 &51.9 &55.7 &62.6 &63.4 &25.9 &37.8 &46.6 &48.9 &51.1 &35.2 &42.9 &47.8 &54.8 &53.5\\
			
			&	Meta FRCN~\cite{metafrcn} & 43.0 & 54.5 &60.6 &\textbf{66.1} &65.4 &27.7 &35.5 &46.1 &47.8 &51.4 &40.6 &46.4 &53.4 &\textbf{59.9} &58.6 \\
			& DeFRCE~\cite{defrcn}                                                                                       & 53.6 & 57.5 & 61.5 & 64.1                   & 60.8                    & 30.1 & 38.1 & 47.0 &  \textbf{53.3}                  & 47.9                    & 48.4 & 50.9 & 52.3 & 54.9                   & 57.4                    \\

			&	FCT~\cite{fct} &49.9 &57.1 &57.9 &63.2 &\textbf{67.1} &27.6 &34.5 &43.7 &49.2 &51.2 &39.5 &54.7 &52.3 &57.0 &58.7\\

			& \textbf{ECEA (Ours)}	&\textbf{59.7} &\textbf{60.7} &\textbf{63.3} &64.1 &64.7 &\textbf{43.1} &\textbf{45.2} &\textbf{49.4} &50.2 &\textbf{51.7} &\textbf{52.3} &\textbf{54.7} &\textbf{58.7} &59.8 &\textbf{61.5} \\
			
			\bottomrule

			& \textit{G-FSOD} \\
			
			&	\multicolumn{1}{l|}{FSRW~\cite{fsrw}}                          &  14.8                      & 15.5                      & 26.7                      & 33.9 & 47.2 & 15.7                      & 15.2                      & 22.7                      & 30.1 & 40.5 & 21.3                      & 25.6                      & 28.4                      & 42.8                      & 45.9                      \\

			&\multicolumn{1}{l|}{Fan \textit{et al.}~\cite{fan}}               &  37.8 &  43.6 & 51.6 & 56.5 & 58.6 & 22.5 & 30.6 & 40.7 & 43.1 & 47.6  & 31.0 & 37.9 &  43.7 & 51.3 & 49.8 \\

			&FSCE~\cite{fsce} &32.9 &44.0 &46.8 &52.9 &59.7 &23.7 &30.6 &38.4 &43.0 &48.5 &22.6 &33.4 &39.5 &47.3 &54.0 \\
			
			&Meta-DETR~\cite{metadetr} &35.1 &49.0 &53.2 &57.4 &62.0 &27.9 &32.3 &38.4 &43.2 &51.8 &34.9 &41.8 &47.1 &54.1 &58.2 \\

			& FCT~\cite{fct} & 38.5 & 49.6 & 53.5 & 59.8 & 64.3 & 25.9 & 34.2 & 40.1 & 44.9 & 47.4 & 34.7 & 43.9 & 49.3 & 53.1 & 56.3 \\
			
			& DeFRCN \cite{defrcn} & 40.2 & 53.6 & 58.2 & 63.6 & \textbf{66.5} & 29.5 & 39.7 & 43.4 & 48.1 & 52.8 & 35.0 & 38.3 & 52.9 & 57.7 & 60.8 \\

			&	\textbf{ECEA (Ours)}                             &     \textbf{55.1}                                           &    \textbf{60.5}                         &       \textbf{62.5}                                                                                                          &   \textbf{63.7}                                                           &    64.0                                                          & \textbf{42.1}                                                         & \multicolumn{1}{c}{\textbf{47.6}} & \multicolumn{1}{c}{\textbf{48.4}} & \multicolumn{1}{c}{\textbf{53.0}} & \multicolumn{1}{c|}{\textbf{57.7}} & \multicolumn{1}{c}{\textbf{39.5}} & \textbf{47.5} & \multicolumn{1}{c}{\textbf{60.7}} & \multicolumn{1}{c}{\textbf{62.8}} & \textbf{66.3}
			\\

			\midrule
			
		\end{tabular}
		
		\label{voc}
	\end{table*}

	\subsection{Model Optimization}
	\label{3.3}
	Faster R-CNN is not directly designed for FSOD. Based on this fact, as shown in Fig. 2, we propose a novel architecture based on Faster R-CNN by ECEA module suitable for few-shot object location recognition. Specifically, the annotation-scarce sample input to the ECEA module generates a series of high-level feature maps. The output feature maps from the ECEA module with the extensible ability are parallelly provided in the next two modules, i.e., RPN and ROI. In addition, under the repeated learning of extensible attention, RPN can get rid of the constraint of anchor fixed size to generate sparse high-quality region proposals sets in the box regression task. Based on sharing extensible part feature vectors and RNP proposals, ROI fine-tunes the box boundaries to contain complete regions of the object. In order to ensure that the detection network will not be dominated by Backbone, RPN, and ROI, we also adopt the decoupled Faster R-CNN strategy \cite{defrcn} to restrict the backward gradient from RPN and ROI to the backbone. Therefore, the loss function from the optimized FSOD architecture can be given by:
	\begin{equation}
	\begin{aligned}
	& \mathcal{L}(x)=\mathcal{L}_{r p n}\left(\left(\mathcal{F}_{r p n}\left(\mathcal{F}_{ecea}(x) ; \theta_{ecea}\right) ; \eta \theta_{r p n}\right), y_{r p n}\right) \\
	&+\lambda \mathcal{L}_{roi}\left(\left(\mathcal{F}_{roi}\left(\left(\mathcal{F}_{ecea}(x) ; \theta_{ecea}\right)\right) ; \gamma \theta_{roi}\right), y_{roi}\right).
	\end{aligned}
	\end{equation}
	Here, $\theta_{ecea}, \theta_{r p n}$, and $\theta_{roi}$ represent learnable parameters for the ECEA, RPN, and ROI module, respectively. $\mathcal{L}_{rpn}$ and $\mathcal{L}_{roi}$ are the loss of RNP and ROI module. $\eta$ and $\gamma$ are decoupling coefficients for RPN and ROI. $\mathcal{F}(\cdot)$  represents learnable function, $y_{rpn}$ and $y_{roi}$ are ground truth, and $\lambda$ is a balanced hyper-parameter for different tasks.

	To implement model optimization, the ECEA module needs to be trained in both the base and novel stages. To be specific, in the base stage, the model fully learns feature extension during sufficient samples to be equipped with the extensible ability which will directly transfer to the novel stage. When some unseen scenarios appear in the novel test sample, the ECEA module can determine whether the appeared regions co-exist with the current object according to extensible learning.

	\begin{table}[t]
		\centering	
		\tabcolsep=0.1cm
		\setlength{\belowcaptionskip}{0.15cm} 
		\caption{
			Performance comparison among ECEA and mainstream  methods based on COCO novel set of 1, 2,  3, 5, 10, and 30 shots. Symbol `-' represents unreported results in the original work.  The results are reported by averaging over multiple runs. Bold font indicates the SOTA result in the group.
		}
		
		\begin{tabular}{ll|cccccc}
			\midrule
			
			&\multicolumn{1}{c|}{\multirow{2}{*}{Methods/shots}} & \multicolumn{6}{c}{Shot Number}       \\
			& \multicolumn{1}{c|}{}                                                                                & 1   & 2   & 3    & 5    & 10   & 30   \\ \midrule
			
			&\textit{G-FSOD}	\\
			& FSRW~\cite{fsrw}                                                                                                 & -   & -   & -    & -    & 5.6  & 9.1  \\
			&	Meta R-CNN~\cite{metarcnn}   & 1.0 & 1.8 & 2.8  & 4.0  & 6.5  & 11.1  \\
			& Meta FRCN~\cite{metafrcn}                                                                                  & 5.1 & 7.6 & 9.8  & 10.8 & 12.7 & 16.6 \\
			
			&FCT~\cite{fct}                                                                                                 & 5.1 & 7.2 & 9.8 & 12.0 & 15.3 & 20.2 \\
			
			& DeFRCN \cite{defrcn}                            & 4.8                                                         & 8.5                         & 10.7                                                         & 13.6                                                         & 16.8                                                         & 21.2                                                         \\
			
			& \textbf{ECEA (Ours)} 	& \textbf{6.1}  & \textbf{10.5}  & \textbf{12.5}  & \textbf{14.9}  & \textbf{18.6}  & \textbf{22.8} \\
			\midrule
			&\textit{FSOD}	\\

			& TFA w/cos~\cite{TFA}                                                                                            & 1.9 & 3.9 & 5.1  & 7.0  & 9.1  & 12.1 \\
			& FSCE~\cite{fsce}                                                                                                 & -   & -   & -    & -    & 11.9 & 16.4 \\
			
			& FCT~\cite{fct}                                                                                                 & 5.6 & 7.9 & 11.1 & 14.0 & 17.1 & 21.4 \\ 
			
			&DeFRCN \cite{defrcn} & 9.3 & 12.9 &14.8 &16.1 &18.5 &22.6 \\

			&\textbf{ECEA (Ours)}&\textbf{9.6} &\textbf{13.2} &\textbf{15.4} &\textbf{16.7} &\textbf{19.6} &\textbf{23.1} \\
			
			\bottomrule
			
		\end{tabular}
		
		\label{COCO}
		
	\end{table}

	\begin{table*}[t]
		\centering
		\caption{
			Performance comparison among ECEA and existing FSOD methods based on all COCO style average precision of 10, and 30 shots. APS, APM, and APL represent AP for small, medium, and large objects, respectively. Bold font indicates the SOTA result in the group. Symbol `-' represents unreported results in the original work. The results are reported by averaging over multiple runs. 
		}
		\begin{tabular}{c|cccccc|cccccc}
			
			\midrule
			\multicolumn{1}{c|}{\multirow{2}{*}{Methods}} & \multicolumn{6}{c|}{10 shots}            & \multicolumn{6}{c}{30 shots}            \\ 
			& AP   & AP50 & AP75 & APS & APM  & APL  & AP   & AP50 & AP75 & APS & APM  & APL  \\ \midrule
			FSRW \cite{fsrw}                    & 5.6  & 12.3 & 4.6  & 0.9 & 3.5  & 10.5 & 9.1  & 19.0 & 7.6  & 0.8 & 4.9  & 16.8 \\
			MetaDet~\cite{metadet}                    & 7.1  & 14.6 & 6.1  & 1.0 & 4.1  & 12.2 & 11.3 & 21.7 & 8.1  & 1.1 & 6.2  & 17.3 \\
			Meta R-CNN~\cite{metarcnn}            & 8.7  & 19.1 & 6.6  & 2.3 & 7.7  & 14.0 & 12.4 & 25.3 & 10.8 & 2.8 & 11.6 & 19.0 \\
			FSDetView \cite{xiao}              & 10.3 & 25.1 & 6.1  & 3.5 & 11.3 & 14.6 & 14.2 & 31.4 & 10.3 & 4.7 & 15.0 & 21.5 \\
			MPSR \cite{sfsod2}                 & 9.8  & 17.9 & 9.7  & 3.3 & 11.3 & 14.6 & 14.1 & 25.4 & 14.2 & 4.0 & 12.9 & 23.0 \\
			TFA   w/fc~\cite{TFA}              & 9.1  & 17.3 & 8.5  & 3.6 & 8.1  & 14.3 & 12.0 & 22.2 & 11.8 & 4.4 & 11.0 & 18.7 \\
			TFA   w/cos~\cite{TFA}             & 9.1  & 17.1 & 8.8  & 3.7 & 8.0  & 14.3 & 12.1 & 22.0 & 12.0 & 4.7 & 10.8 & 18.6 \\
			FSCE~\cite{fsce}                    & 11.4 & 23.3 & 10.1 & 4.5 & 10.8 & 18.7 & 15.8 & 29.9 & 14.7 & 6.1 & 10.8 & 18.6 \\
			SRR-FSD \cite{sfsod3}                & 11.3 & 23.0 & 9.8  & -    &  -    &  -    & 14.7 & 29.2 & 13.5 & -    & -     & -     \\
			PTF \cite{tip1}                    & 11.7 & 22.6 & 10.9 & 5.1 & 12.2 & 16.5 & 15.7 & 30.4 & 14.4 & 7.3 & 16.3 & 21.4 \\
			PTF+KI \cite{tip1}                 & 13.0 & 24.0 & 12.6 & 6.0 & 13.1 & 18.4 & 16.8 & 30.9 & 16.2 & 8.0 & 17.1 & 23.0 \\
			CKPC \cite{tip2}                   & 16.6 & 34.4 & 17.2 & 5.9 & 18.3 & 27.5 & 19.9 & 38.1 & 19.7 & 7.8 & 20.9 & 31.3 \\
			\textbf{ECEA (G-FSOD)}             & 18.6 & 33.4 & 18.4 & 7.3 & 18.1 & 29.2 & 22.8 & 39.0 & 23.3 & 8.9 & 23.0 & \textbf{35.3} \\
			\textbf{ECEA (FSOD)}              & \textbf{19.6} & \textbf{34.6} & \textbf{19.2} & \textbf{8.6} & \textbf{19.6} & \textbf{29.3} & \textbf{23.1} & \textbf{39.5} & \textbf{23.8} & \textbf{9.7} & \textbf{23.1} & 34.6 \\
			\midrule
		\end{tabular}
		\label{COCOapall}
	\end{table*}

	\begin{table}[t] 
		\centering

		\setlength{\belowcaptionskip}{0.1cm} 
		\caption{
			Performance comparison among ECEA and mainstream methods on PASCAL VOC and COCO base set. Symbol `*' indicates that the results are reproduced by us. Symbol `-' represents unreported results. Bold font indicates the SOTA result in the group.
		}

		\begin{tabular}{cc|ccc}
			\hline
			\multicolumn{2}{c|}{Dataset / methods}                                 & \multicolumn{1}{c}{TFA} & \multicolumn{1}{c}{DeFRCN} & \multicolumn{1}{c}{\textbf{ECEA}} \\ \hline
			\multicolumn{1}{c|}{\multirow{4}{*}{bAP50}} & Base1                    & 80.8                     & 80.3                        & \textbf{82.1}                     \\
			\multicolumn{1}{c|}{}                       & Base2                    & 81.2*                     & 81.7                        & \textbf{82.4}                     \\
			\multicolumn{1}{c|}{}                       & Base3                    & 81.4*                     & 81.1                        & \textbf{82.8}                    \\
			\multicolumn{1}{c|}{}                       & COCO                     &    -                      & 59.2                            &       \textbf{60.2}                    \\ \midrule
			\multicolumn{1}{c|}{\multirow{4}{*}{bAP75}} & Base1                    & 59.3*                     & 60.2                        & \textbf{62.3}                     \\
			\multicolumn{1}{c|}{}                       & Base2                    & 60.9*                     & 61.5                        & \textbf{62.6}                     \\
			\multicolumn{1}{c|}{}                       & Base3                    & 61.2*                     & 60.9                        & \textbf{63.9}                     \\
			\multicolumn{1}{c|}{}                       & \multicolumn{1}{c|}{COCO} &   \multicolumn{1}{c}{-}     & \multicolumn{1}{c}{42.2}        & \multicolumn{1}{c}{\textbf{42.8}}     \\ \midrule
		\end{tabular}
		
		\label{base}
	\end{table}

	\section{Experimental Results and Analysis}
	
	In this section, we first introduce the experimental settings in section \ref{4.1}. We then verify the performance of our proposed ECEA module compared with multiple classical works on the PASCAL VOC~\cite{voc} and COCO~\cite{coco} datasets in section \ref{4.2}. Finally, we design comprehensive ablation studies on the proposed model architecture in section \ref{4.3}.
	
	\subsection{Experimental Setting}
	\label{4.1}
	\textbf{Dataset Setting.} Following previous benchmarks~\cite{TFA, fsce, defrcn}, we use the PASCAL VOC and COCO datasets to verify the performance of our method. As for PASCAL VOC, we combine VOC2007 and VOC2012 to train the proposed network, in which the combined dataset is split into three groups same as \cite{TFA,fsce}. To be specific, each group in the dataset includes 15 base classes and 5 novel classes, where each novel class is randomly combined with 1, 2, 3, 5, and 10 shots for reporting accuracy. Furthermore, the VOC2007 test set is used for evaluation. With regard to COCO, 60 categories disjoint with VOC are marked as base classes, and meanwhile, the remaining 20 classes are used as novel classes. We then select 5,000 images from the val dataset for testing and the rest combined with the training set for training. We follow the validation strategy of the previous benchmarks ~\cite{fct,defrcn} to report the detection accuracy under random shots 1, 2, 3, 5, 10, and 30. 
	
	\textbf{Evaluation Setting.} We consider two evaluation protocols, i.e., FSOD and G-FOSD, to access the effectiveness of our approach. Specifically, FSOD only focuses on the performance of novel classes. In G-FSOD, considering both base and novel classes not only accesses the performance of the novel classes but also can monitor the catastrophic forgetting of the base classes ~\cite{retina,cfa}, which reports the overall performance more comprehensively than FSOD. As for evaluation metrics, we report AP50 (matching threshold is 0.5) on PASCAL VOC and COCO style mean average precision (mAP). In addition, the evaluation protocol of all reported results is averaged over multiple repeated runs.
	
	\textbf{Implementation Details.} We adopt Faster R-CNN~\cite{fasterrcnn} as the basic detection framework, ResNet101~\cite{res101} pre-trained on ImageNet~\cite{imagenet} as the backbone. Furthermore, all experiments use SGD to optimize our proposed model with the momentum of 0.9 and weight decay of $5e^{-5}$, in which the learning rate is set to 0.015 with the batch size of 6 based on one GPU (RTX 3090 Ti).

	\subsection{Comparison Results}
	
	\label{4.2}

	\subsubsection{Results on PASCAL VOC}
	We report the novel AP50 (nAP50) of three random novel splits on PASCAL VOC. Table \ref{voc} lists the results of ECEA compared with existing classical methods in G-FSOD and FSOD settings, respectively. From the table, our approach achieves satisfactory performance improvement over both protocols, where 11 group shots reach the best results in FSOD and 14 groups in G-FSOD among 15 group results. In addition, ECEA as a transfer-learning-based method with a simple design concept achieves better performance than FCT~\cite{fct} as a meta-learning-based approach. For example, compared with FCT, 14 groups results achieve better results in the FSOD setting. The results on VOC also verify that the performance of the model can be improved by extending co-existing regions of few-shot objects.

	\begin{figure}[t]
		\vspace{-0.1cm}
		\centering

		\includegraphics[width=9cm,height=5.73cm]{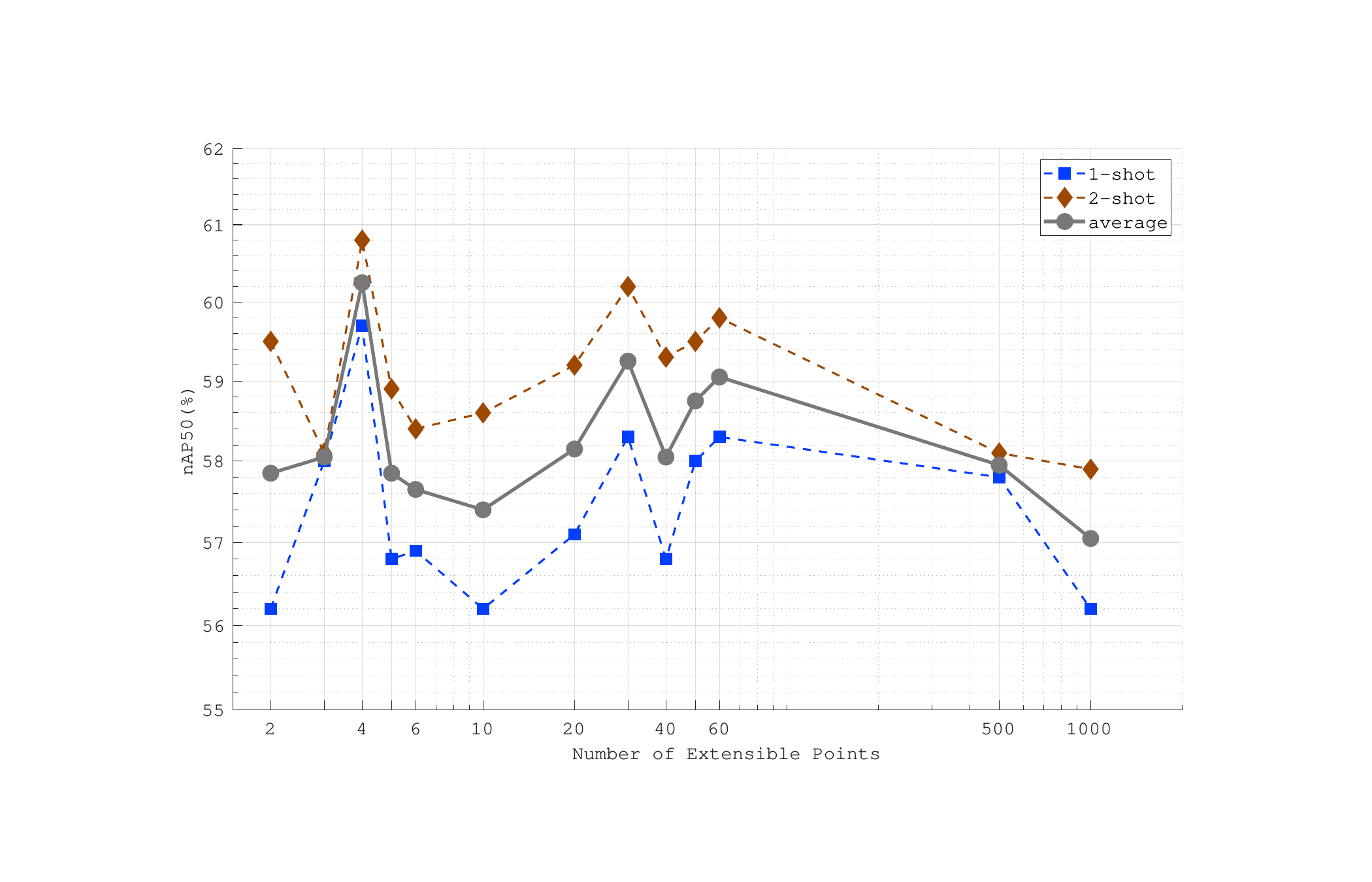}  
		\vspace{-0.3cm}
		\caption{Performance comparison among various extensible points setting. }
		
		\label{f4}

	\end{figure}

	\begin{figure}[t]
		\vspace{-0.1cm}
		\centering
		\includegraphics[width=8.08cm,height=8.5cm]{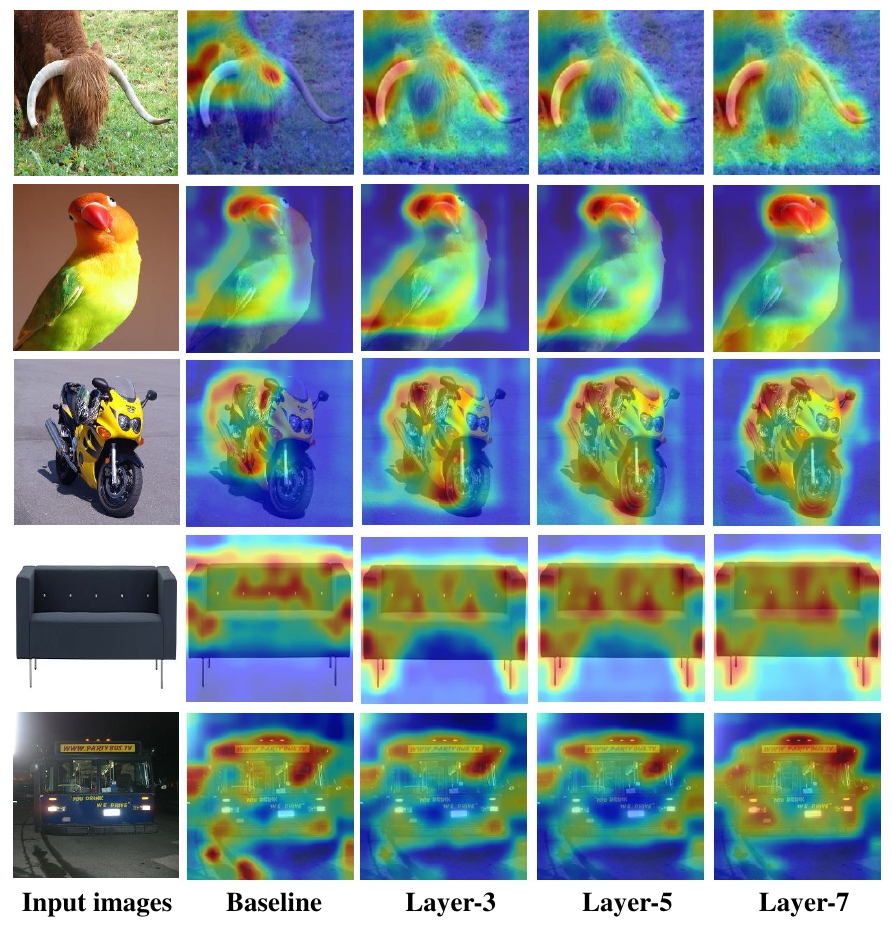} 
		\vspace{-0.1cm}
		\caption{Performance of localization in different extensible attention layers. Layer-3, Layer-5, and Layer-7 are the results in 7 extensible attention multi-layers. Baseline represents the model without extensible attention.}
		
		\label{f6}
		\vspace{-0.1cm}
	\end{figure}

	\begin{figure*}[t]
		\vspace{-0.1cm}
		\centering
		\includegraphics[width=17cm,height=6.37cm]{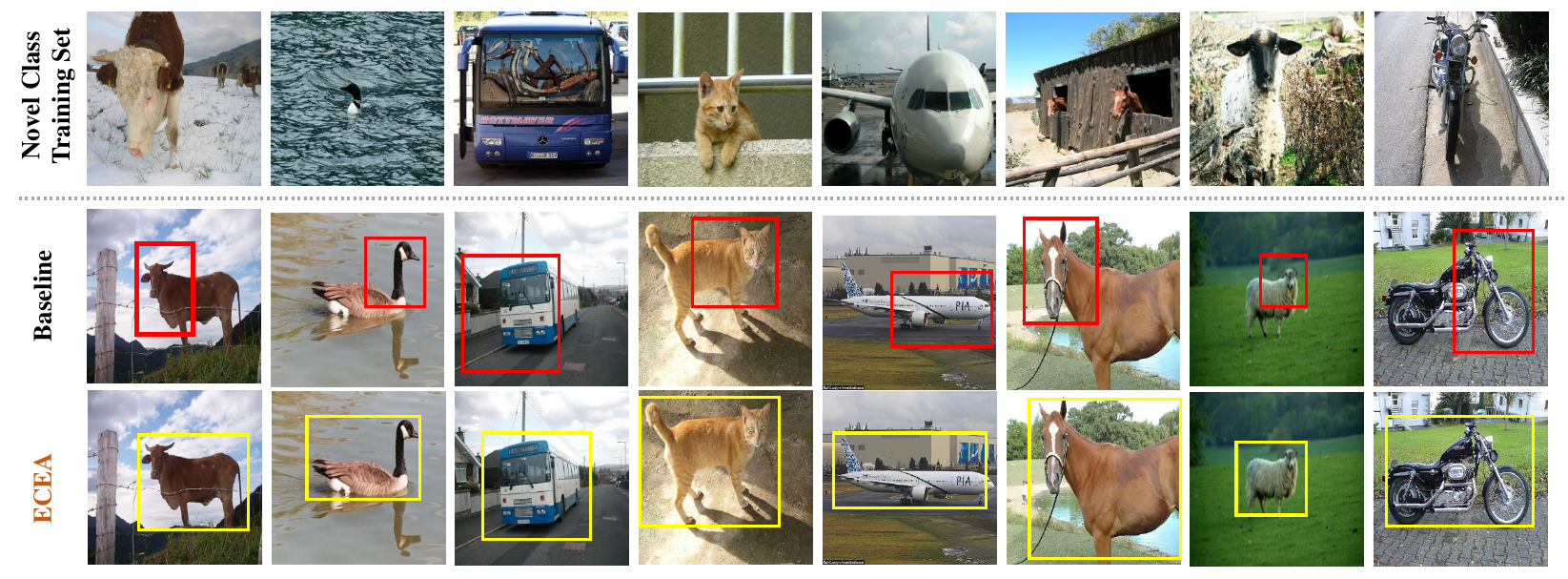} 
		\vspace{-0.1cm}
		\caption{ Prediction of the single object in an image on the VOC2007 test dataset. The baseline FSOD method, limited by the number of training samples, may only identify part features of the object. In contrast, our ECEA can enable the model to infer the global object according to the local parts.
		}

		\label{f5}
		\vspace{-0.1cm}
	\end{figure*}

	\subsubsection{Results on COCO} The COCO dataset covers more novel categories than VOC and has more complex sample scenarios, which can better illustrate the generalization ability of various FSOD methods. Table \ref{COCO} shows the comparison results of recent classical FSOD approaches and ECEA on the COCO dataset. From the table, all evaluation results of ECEA outperform the classical methods in G-FSOD and FSOD protocols, respectively. In addition, ECEA at FSOD protocol achieves 10.4\% and 10.6\% improvement over the baseline of fine-tuning-based TFA~\cite{TFA} in 10-shot and 30-shot settings, respectively. And meanwhile, compared with FCT~\cite{fct} based on meta-learning and transformer, our approach achieves 2.5\% and 1.7\% improvement, respectively. All results on COCO indicate that the model with the ECEA module is equipped with strong extension and generalization ability in FSOD.
	
	In addition, to further verify the performance of our proposed method, Table \ref{COCOapall} lists the results, based on all COCO style average precision, of comparison among ECEA and existing FSOD methods of 10-shot and 30-shot settings. As can be seen from the table, our method achieves significant improvement in AP75 (the matching threshold is 0.75). In 10-shot setting, compared with the latest FSOD study \cite{tip2}, the performance of ECEA is enhanced by 1.2 and 2 points in both G-FSOD and FSOD protocols, respectively. And meanwhile, 3.6\% and 4.1\% improvements in both G-FSOD and FSOD protocols are achieved over the CKPC \cite{tip2} in 30 shot setting. The results fully demonstrate that our proposed ECEA can effectively assist the few-shot model to extend global co-existing regions from the local part. From all comparison results, our method significantly improves the performance of the few-shot model in both FSOD and G-FSOD settings. In the FSOD protocol, our method is higher than existing FSOD methods in all indicators. It can be verified that enabling the model to infer the global object according to the local parts can significantly improve the performance of the few-shot model.

	\subsubsection{Effectiveness of ECEA Module on Base Classes}
	We conjecture that the ECEA module may improve the base model through extensible learning. To verify this speculation, based on VOC and COCO datasets, we evaluate the performance of ECEA compared with two classical works on base categories. Table \ref{base} illustrates the results of base classes on AP50 and AP75. As shown in the table, our ECEA is superior to the previous works~\cite{TFA,defrcn} in both of AP50 and AP75. In particular, our approach achieves SOAT performance in object regression, e.g., 3.0\% and 2.7\% improvements in base AP75 are achieved over the classical DeFRCN and TFA on VOC base3, respectively. Therefore, the ECEA module is as well as effective in data-sufficient scenarios, which also indicates that our method may have a wider application prospect.

	\begin{table}[t]
		\centering
		\tabcolsep=0.09cm
		\tabcolsep=0.1cm
		\renewcommand{\arraystretch}{1.2}
		\setlength{\belowcaptionskip}{0.1cm}
		\caption{
			Comparison of extensible attention with different ResNet101 stages. Avg represents the average results of five group of shots. Bold font indicates the SOTA result in the group.	}

		\begin{tabular}{lccc|cccccc}
			\hline

			\multicolumn{4}{c|}{Stages}  & \multicolumn{6}{c}{Shot Number}                                                                                           \\  
			& Block3 & Block4 & Block5 & 1        & 2        & 3        & 5        & \multicolumn{1}{c}{10}       & Avg       \\  \hline

			& \CheckmarkBold  &    &    & 15.1          & 17.7          & 15.1          & 20.9          & 18.7          & 17.5          \\ 
			&    & \CheckmarkBold  &    & 37.1          & 42.5          & 39.7          & 49.5          & 49.9          & 43.7          \\ 
			&    &    & \CheckmarkBold  & 41.5          & 45.5          & 49.0          & 58.4          & 52.4          & 49.4          \\ 
			&    & \CheckmarkBold  & \CheckmarkBold  & 47.4          & \textbf{51.6} & 52.1          & 63.3          & 60.8          & 55.0          \\ 
			& \CheckmarkBold  & \CheckmarkBold  & \CheckmarkBold  & \textbf{47.5} & 49.4          & \textbf{53.3} & \textbf{64.8} & \textbf{61.3} & \textbf{55.3} \\
			\hline
		\end{tabular}

		\label{block}
	\end{table}

	\subsection{Ablation Study}
	
	\label{4.3}

	\subsubsection{Combination Setting of Extensible Attention with ResNet101 Hierarchies}
	In the ECEA module, extensible attention can assist the few-shot detector to automatically capture co-existing regions in the feature map, while the residual module in ResNet101 can effectively solve the problem of gradient disappearance in deep networks to improve model accuracy. The appropriate combination of these two approaches in different feature scales can progressively discover the full object in various receptive fields and significantly enhance the generalization ability of the model. Therefore, we perform comprehensive ablation experiments on VOC-Split1 to explore the most reasonable combination of the different output stages of ResNet101 and extensible attention. Furthermore, we refer to the combination of a stage and extensible attention as a block in this paper. As for the training details, the number of extensible points is set to 4 and the number of multi-layers of extensible attention is set to 7. Notably, we remove the decoupling module to release a certain GPU memory to report the base results in the G-FSOD setting.
	
	Table \ref{block} lists the experiment results for the three different blocks, where a block is the combination of a ResNet101 stage with extensible attention. As can be seen from the table, Block5 learns more high-level features to obtain the best scalability and accuracy in the single block output. We thus always consider Block5 to combine with other blocks. From the table, Block5 with Block3 and Block4 reach the best results. In view of this, we select the last three stages of ResNet101 combined with the extensible attention in the ECEA module.

	\begin{table}[t]
		\centering
		\tabcolsep=0.2cm
		\vspace{0.5cm}
		\caption{
			Performance of different extensible attention layers. Bold font indicates the SOTA result in the group.}

		\begin{tabular}{c|cccccc}
			\cline{1-7}
			Extensible- & \multicolumn{6}{c}{Shot Number}                                                                                           \\ 
			Layers & 1             & 2             & 3             & 5            & \multicolumn{1}{c}{10}            & \multicolumn{1}{c}{Avg} \\ \hline 
			2                           & 57.3          & 58.6          & 60.4          & 62.8          & \multicolumn{1}{c}{62.8}          & 60.4                         \\  
			3                           & 59.3   & 59.8          & 60.1          & 61.6          & \multicolumn{1}{c}{63.2}          & 60.8                         \\  
			4                           & 57.8          & 59.3          & 60.2          & 63.8    & \multicolumn{1}{c}{63.6}          & 60.9                         \\  
			5                           & 57.6          & 58.5          & 59.0          & \textbf{65.0} & \multicolumn{1}{c}{63.2}          & 60.7                         \\  
			6                           & 57.7          & 58.4          & 59.9          & 61.7          & \multicolumn{1}{c}{59.6}          & 59.5                         \\  
			7                           & \textbf{59.7} & \textbf{60.7} & \textbf{63.3}    & 64.1    & \multicolumn{1}{c}{\textbf{64.7}} & \textbf{62.5}                \\ 
			8                           & 58.5          & 59.6          & 61.0          & 60.2          & \multicolumn{1}{c}{60.6}          & 60.0                         \\  
			10                          & 59.3          & 60.5    & 62.5 & 62.0          & \multicolumn{1}{c}{64.0}    & 61.7                   \\ \hline 
		\end{tabular}

		\label{layers}

	\end{table}

	\begin{figure*}[t]
		\vspace{-0.1cm}
		\centering
		\includegraphics[width=17cm,height=10.3cm]{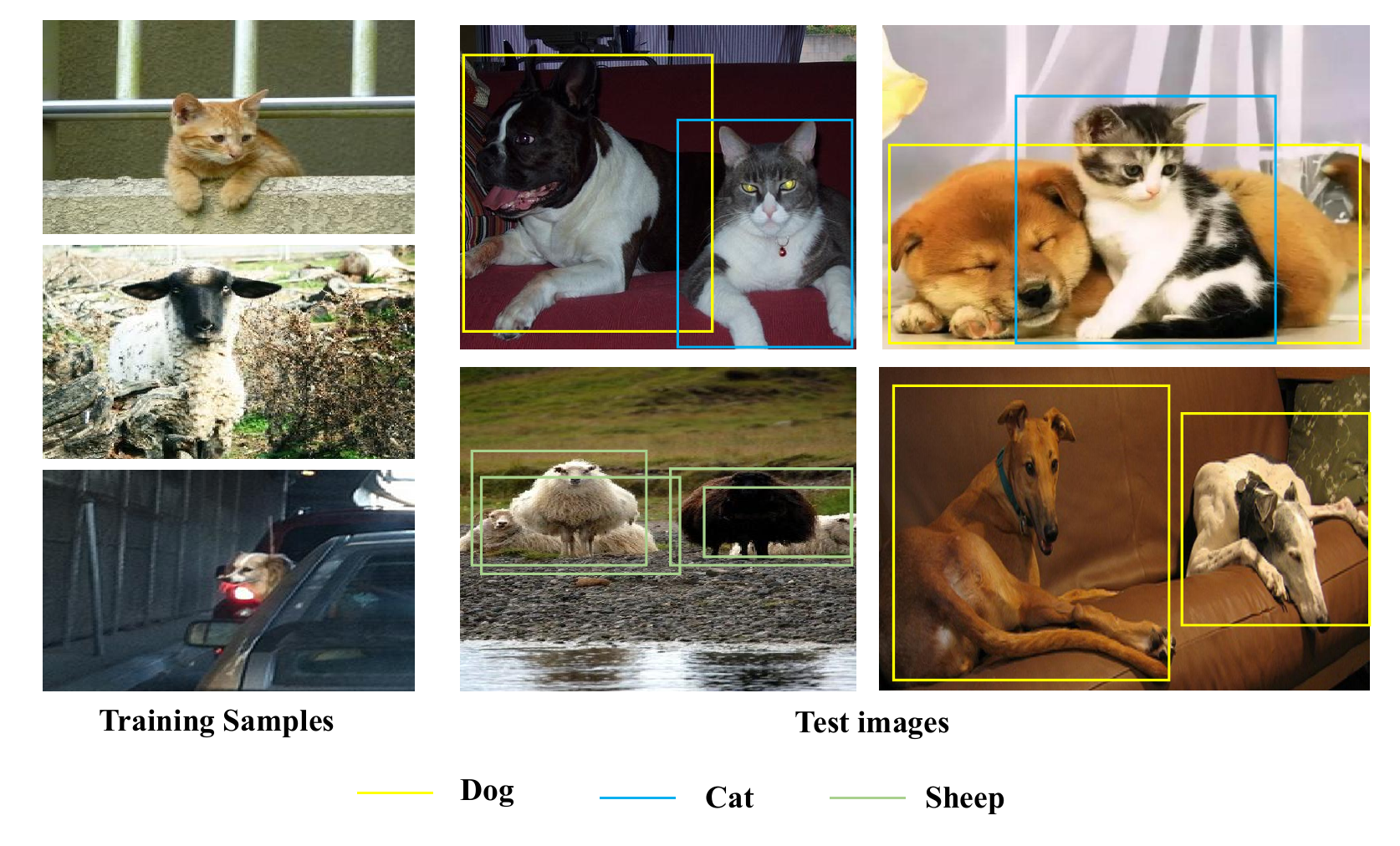}
		\vspace{-0.1cm}
		\caption{ Prediction results of ECEA in an image with multiple occlusion objects. Since the first row belongs to different objects, each category can be accurately predicted. Since the objects of the first image in the second row are the same, the prediction bounding boxes range is larger, but no object feature is missed. Secondly, it could correctly predict the two dogs without occluding each other.}
		\label{f7}
		\vspace{-0.1cm}
	\end{figure*}
	
	\subsubsection{Extensible Points Setting} In extensible attention, a certain sample location with increasing key points can extend more prominent regions, but the computational complexity also grows. To ensure that the performance of ECEA with the few attention calculations would not decline, we compare the results between a series of groups of extensible points on VOC-Split1. In this experiment, we take the results of 1-shot, 2-shot, and the average as our evaluation index. To be specific, Fig. \ref{f4} shows the results of 13 groups of extensible points. There are three groups of local maximum extensible points in the figure, which are 4, 30, and 50, respectively. The first point is also the absolute maximum point. On the other hand, as the number of points increases, a lot of computational complexity would be introduced, which consumes more training time than the former setting, especially in over 1000 points. According to the above analysis, we thus consider that a sample point with 4 prominent extensible regions is most suitable for the ECEA module. Meanwhile, the computational complexity of extensible attention can be expressed as $O(4HWC)$.
	
	It is worth noting that as the number of extensible points in our method increases, the effect is closer to the normal transformer \cite{selfattention}, and when the extensible points are equal to the pixels of the entire image, the effect is equal to the normal transformer.

	\subsubsection{Effectiveness of Multi-Layer Extensible Attention} With the increase of attention layers, the model can learn feature extension repeatedly, but also enhance the network complexity which would lead to overfitting. We evaluate the performance of the model at different levels through a series of experiments on the VOC-Split1 set. Table \ref{layers} lists the results for the 8 groups of layers. From the table, the 7 multi-layer reach the highest in the average results, and in the 5 groups of shot results, three groups reach the SOTA, and two groups rank second.
	
	Furthermore, we have supplemented the running time of each additional extensible attention layer in the paper. Specifically, by counting the average running time of 4952 test images on VOC, we found that each additional layer of layer extensible attention consumes about 0.687 milliseconds. while the average inference time per image is 50 milliseconds without any extensible attention layer. Therefore, the increased time consumption can be negligible.
	
	In addition, Fig. \ref{f6} shows the performance of the model extends objects in three extensible attention layers on VOC-Split1 classes. We find an interesting phenomenon in this figure that our approach can not only scale out from the part feature to find relevant features, but also scale in from misidentified features in the baseline to gradually eliminate redundant features. For example, with the increase of extensible attention layers, the cow, motorbike, bird, and sofa are gradually recognized, completely. On the contrary, the bus and class-unrelated features are together identified in the baseline. With the increase of extensible attention layers, redundant features are gradually removed.

	\subsection{One Object in the Test Image} 
	\label{4.4}
	Processing with a single-part object can effectively extend co-existing parts in an image by our proposed method. For example, Fig. \ref{f5} shows the results of one object in the test image detection of ECEA with the 1-shot setting on VOC. The novel training samples are randomly selected, in which the results are detected by the best training models of three splits. From the figure, each bounding box can completely cover the corresponding object in ECEA results. 
	
	\subsection{Multiple Objects in the Test Image }
	\label{4.5}
	When the part object is blocked by the same category in an image, our ECEA can assist models avoid missing feature recognition. The advantage of our method lies in solving incomplete object feature prediction, but might sometimes scale the same objects into a single object which is also a general problem in crowded object detection \cite{crowdob}. However, if a part object is blocked by the other novel object in an image, our method can assist the model to completely and correctly predict them due to the given features of objects being different. Our method can effectively extend the co-existing features in different objects in the above scenarios. Fig. \ref{f7} as an example lists the prediction results of ECEA in an image with multiple occlusion objects.

	\vspace{-0.1cm}

	\section{Discussion}
	
	As discussed, the existing FSOD approaches seldom consider the localization of objects from local to global. Limited by the scarce training data in FSOD, novel classes are highly possible to provide training samples that only capture part of objects, resulting in such methods cannot detect the complete object during testing. We consider that the model can infer the complete object from the abundant co-existing part features in the base stage. We thus design the ECEA module to assist the model learn the extensible ability from local to global and transfer it from base classes to novel classes, which can assist the few-shot model to completely detect the object during testing.
	
	We take extensive experiments to support our motivation and demonstrate the effectiveness of our ECEA module. Section \ref{4.2} presents the comparisons with state-of-the-art methods both in G-FSOD and FSOD settings, which shows that ECEA can achieve state-of-the-art performance compared to the existing transfer-learning-based and meta-learning-based approaches. We further conduct ablation studies in Section \ref{4.3} to verify the effects of each component of ECEA. Furthermore, in Section \ref{4.4} and \ref{4.5}, we provide the results of one object and multiple objects in a test image, respectively, to intuitively demonstrate that ECEA can assist the FSOD model to effectively extend the co-existing features from local to global.
	
	Indeed, the ECEA module is limited in this condition. When the part object is blocked by the same category in an image, our ECEA can assist models avoid missing feature recognition, but might sometimes scale the same objects into a single object. 
	
	Fortunately, the proposal-based object detector \cite{setnms} equipped with EMD Loss and Set NMS lets each proposal predict a set of correlated instances rather than a single one to effectively detect highly overlapped objects. Adopting this proposal can alleviate the limitation of ECEA.

	\section{Conclusion}
	
	In this paper, we have investigated an FSOD method, namely the ECEA module, which alleviates the data scarcity problem for FSOD to detect unseen object parts. Specifically, we proposed the extensible attention mechanism which lets the model from a current region infer all co-existing regions of the common object. Furthermore, we combined the feature extractor with the extensible attention mechanism at different feature scales in the ECEA module, which progressively discovered the full object in various receptive fields. Based on the PASCAL VOC and COCO datasets, we have taken extensive comparison and ablation experiments to evaluate the effectiveness and expandability of our method, respectively. The experiment results show that the ECEA module could completely detect unseen object parts and achieved new states of the arts in FSOD.

	\bibliographystyle{IEEEtran}
	\bibliography{PAM}

	\ifCLASSOPTIONcaptionsoff
	\newpage
	\fi

\end{document}